\documentclass{article} % For LaTeX2e
\usepackage{iclr2026_conference,times}

\usepackage{amsmath,amsfonts,bm}

\def\eqref#1{equation~\ref{#1}}
\def\1{\bm{1}}

\DeclareMathAlphabet{\mathsfit}{\encodingdefault}{\sfdefault}{m}{sl}
\SetMathAlphabet{\mathsfit}{bold}{\encodingdefault}{\sfdefault}{bx}{n}

\usepackage{hyperref}
\usepackage{url}
\usepackage{amsmath,amsfonts,amssymb,amsthm,enumerate,listings,appendix,xcolor,tikz,MnSymbol}
\usepackage{microtype,booktabs,mathtools,cleveref,siunitx,algorithm,algorithmic,graphicx,multirow}
\usetikzlibrary{decorations.pathreplacing,calligraphy}

\newcommand{\hard}{\textsc{Hard}}
\newcommand{\easy}{\textsc{Easy}}
\newcommand{\midsel}{\textsc{Medium}}
\newcommand{\rand}{\textsc{Random}}

\title{Hard Examples Are All You Need: Maximizing GRPO Post-Training Under Annotation Budgets}

\author{
Benjamin Pikus\thanks{Equal contribution}\\
Independent\\
\And
Pratyush Ranjan Tiwari\footnotemark[1]\\
Eternis Labs\\
\And
Burton Ye\\
Writer, Inc\\
}

\iclrfinalcopy % Uncomment for camera-ready version, but NOT for submission.
\begin{document}

\maketitle

\begin{abstract}
Collecting high-quality training examples for language model fine-tuning is expensive, with practical budgets limiting the amount of data that can be procured.
We investigate whether example difficulty affects GRPO training effectiveness by comparing selection strategies (easy, medium, hard, random) across multiple models and reasoning tasks.
Training on the hardest 10\% of examples (those where the base model fails most often) yields dramatic performance gains up to 47\%, while easy examples produce minimal improvements of 3-15\%.
This occurs because GRPO requires outcome variance to generate learning signals; hard examples maintain mixed success/failure outcomes throughout training while easy examples quickly converge to consistent success, eliminating learning opportunities. Moreover, models trained on hard examples show superior out-of-distribution generalization, with only hard-trained models achieving meaningful gains on the AIME2025 benchmark. Our findings provide clear guidance: when budget-constrained, prioritize collecting and annotating examples where your base model struggles, as these drive nearly all learning value in GRPO fine-tuning.
\end{abstract}

\section{Introduction}

Large language model (LLM) alignment and post-training are fundamentally constrained by the expense of acquiring high-quality supervision data, making the selection of training examples a critical factor in achieving optimal model performance \cite{Ouyang2022InstructGPT}.\footnote{Code available \href{https://anonymous.4open.science/r/grpo_difficulty-BF42/README.md}{here}}
While recent advances in reinforcement learning from human feedback (RLHF) and related techniques have demonstrated remarkable improvements in model capabilities, practitioners face a practical challenge: given limited resources for data annotation and curation, which examples should be prioritized to maximize post-training performance?

In this work, we focus on Group Relative Policy Optimization (GRPO), a PPO‑style algorithm that replaces a learned value function with group‑normalized advantages, reducing memory and relying on within‑group reward variance for learning signals. We address a specific instantiation of this budget-aware selection problem: given a fixed budget to select and train on only a fraction of available prompts, how should we choose this subset to maximize the effectiveness of Group Relative Policy Optimization (GRPO) fine-tuning? 
Specifically, we investigate whether selecting examples based on their difficulty (as measured by the base model's success rate across multiple sampling attempts) leads to systematic differences in final model performance on held-out test sets.

Our key research questions are:
\begin{enumerate}
\item Should practitioners prioritize examples that are hard (where the base model frequently fails), easy (where it frequently succeeds), medium, or a selection of random difficulty examples?
\item Does the optimal selection strategy vary across model scales and families?
\item What mechanisms explain the differential effectiveness of difficulty-based selection strategies under GRPO's learning dynamics?
\item How do different difficulty-based selection strategies impact out-of-distribution generalization to substantially harder problem sets?
\item Can we achieve similar performance gains by training only on examples the base model gets wrong, avoiding the circular dependency of knowing which examples will improve?
\end{enumerate}

Our findings reveal a striking pattern: \textbf{hard examples are all you need}. Training on the hardest 10\% of examples yields performance gains up to 47\%, while easy examples produce minimal improvements of 3-15\%. This 30+ percentage point gap transforms marginally effective fine-tuning into highly successful model improvement.

To answer these questions, we make the following contributions:

\medskip \noindent \textbf{Budget-aware evaluation protocol.} We develop a systematic framework for comparing difficulty-conditioned training subsets under GRPO, using multi-sample base-model probing to robustly estimate example difficulty \cite{Wang2022SelfConsistency}.
This protocol enables fair comparison across selection strategies while controlling for computational budget and other confounding factors.

\medskip \noindent \textbf{Comprehensive experimental evaluation.} We conduct extensive experiments on GSM8K grade-school math problems and BIG-Bench Hard's Tracking Shuffled Objects task across models (Qwen3-4B, Qwen3-14B, Phi-4, and Llama3.1-8B) \cite{dubey2024llama, qwen3technicalreport, phi4technicalreport}.
Our results consistently show that training on the hardest 10\% of examples yields superior test performance on reasoning tasks.

\medskip \noindent \textbf{Mechanistic understanding of selection effects.} We analyze why hard examples prove most effective under GRPO: the algorithm requires within-group outcome variance to generate learning signals.
When all samples in a group produce identical rewards, the advantages become zero and learning stops.
Our analysis shows that hard examples maintain mixed outcomes longer than easy examples, which quickly converge to deterministic success \cite{Shao2024DeepSeekMath}.

\medskip \noindent \textbf{Out-of-distribution generalization analysis.} We evaluate models trained on different difficulty-based subsets against the AIME2025-I benchmark, a substantially harder test set than the in-distribution GSM8K \cite{opencompass_AIME2025, matharena}.
We find that training on the hardest examples not only improves in-distribution performance, but is also the only strategy that yields meaningful gains on OOD data.

\medskip \noindent \textbf{Base wrong vs base right analysis.} We validate our ``hard examples'' principle from a complementary perspective by categorizing examples based on base model performance. This analysis reveals that nearly all learning value comes from examples the base model initially gets wrong, with training on ``base wrong'' examples achieving up to 23.5\% relative improvements over ``base right'' examples. This convergent evidence that both low pass@k examples and base-wrong examples drive learning confirms that GRPO fundamentally benefits from training at the frontier of model capabilities.

%\textbf{Fine-grained behavioral analysis.} We analyze what specific capabilities hard-trained models acquire that others miss, identifying systematic improvements in multi-step reasoning, error recovery, and compositional generalization. This analysis provides actionable insights into the types of examples that drive meaningful capability improvements.

Our findings have immediate practical implications: when using GRPO for reasoning task fine-tuning under budget constraints, practitioners should prioritize collecting and annotating examples where the base model struggles. 
We validate this principle from two complementary angles: examples with low pass@k success rates and examples the base model gets wrong both drive superior learning outcomes.
This convergence demonstrates that ``hard examples'' identified through either metric capture the same underlying phenomenon: problems at the frontier of model capabilities that provide rich learning signals.
This hard-example focus can yield performance improvements \textit{exceeding 30 percentage points} over easy selection on some configurations, transforming marginally effective fine-tuning into highly successful model improvement.

Our study isolates \emph{offline} difficulty‑based selection under GRPO. It complements \emph{online} difficulty‑targeted selection with rollout replay that prioritizes moderate difficulty \cite{Sun2025DiffSelReplayGRPO}, advances in process‑reward RL (PRIME) \cite{Cui2025PRIME}, and analyses of GRPO optimization biases such as Dr.\ GRPO \cite{Liu2025DrGRPO}. We focus on outcome‑reward GRPO; extending our protocol to dense process rewards or debiased GRPO is left to future work.

The remainder of this paper is organized as follows. Section 2 presents our experimental protocol for difficulty estimation and subset selection.
Section 3 reports our main experimental results across tasks and model scales.
Section 4 provides detailed analysis of learning dynamics and model behavior.
Section 5 concludes with practical recommendations and future directions. Related work can be found in the appendix.

\section{Experimental Protocol}

This section describes our experimental framework for evaluating the impact of example difficulty on GRPO fine-tuning under budget constraints.
We detail our difficulty estimation procedure, subset selection policies, training methodology, and evaluation metrics.

\subsection{Goal}

Our primary research goal is to determine which subset selection strategy maximizes post-GRPO test accuracy when constrained to using only $p = 10\%$ of the available training pool.
Specifically, given a pool of unlabeled prompts $\mathcal{X}$, we select subsets $S \subset \mathcal{X}$ with $|S| = \lfloor p|\mathcal{X}| \rfloor$ according to different difficulty-based policies, train models using GRPO on each subset, and compare their performance on held-out test sets.
This setup mirrors practical scenarios where annotation budgets limit the number of examples that can be used for fine-tuning.

\subsection{Difficulty estimation via multi-sample probing}
\label{sec:difficulty}

To robustly estimate example difficulty, we employ multi-sample evaluation using the base model before any fine-tuning.
For each prompt $x$ in the training pool, we sample $K$ independent completions from the base model $\pi_{\text{base}}$ using temperature $\tau = 1.0$ and chain-of-thought prompting where applicable \cite{Wang2022SelfConsistency}. For each prompt $x$, we compute:
\begin{align}
\hat{p}(x) &= \frac{1}{K} \sum_{i=1}^{K} \mathbf{1}[\text{completion}_i \text{ is correct}] %\\
%s^2(x) &= \hat{p}(x)(1 - \hat{p}(x))
\end{align}

where $\hat{p}(x)$ represents the empirical success rate.

\medskip \noindent \textbf{Task-specific correctness criteria:}
\begin{itemize}
\item \textbf{GSM8K:} Exact match of the final numerical answer after extracting from the generated solution \cite{Cobbe2021GSM8K}
\item \textbf{Tracking Shuffled Objects:} Correct identification of all object positions after the described swaps \cite{Suzgun2023BBH}
%\item \textbf{KEGG knowledge task:} Selection of the correct multiple-choice option
\end{itemize}

We use $K = 10$ samples for Tracking Shuffle Objects, and $K = 5$ for GSM8K due to its larger size and associated computational costs.
This multi-sample approach provides stable difficulty estimates while remaining computationally tractable.

\subsection{Subset selection policies}

Given difficulty estimates for all prompts, we implement four selection policies, each choosing exactly $10\%$ of the training pool:

\begin{enumerate}
\item \textbf{\hard{}(-est):} Select prompts with the lowest success rates:
\[
S_{\text{hard}} = \arg\min_{S: |S|=\lfloor p|\mathcal{X}| \rfloor} \sum_{x \in S} \hat{p}(x)
\]

\item \textbf{\easy{}(-est):} Select prompts with the highest success rates:
\[
S_{\text{easy}} = \arg\max_{S: |S|=\lfloor p|\mathcal{X}| \rfloor} \sum_{x \in S} \hat{p}(x)
\]

\item \textbf{\midsel{}:} Select prompts nearest to the median difficulty, specifically those in the interquartile range around the median $\hat{p}(x)$ value

\item \textbf{\rand{}:} Uniform random sample without replacement from the full pool
\end{enumerate}

Figure~\ref{fig:pipeline} illustrates the complete pipeline from difficulty estimation through subset selection to GRPO training.

\begin{figure}[h]
\centering
\fbox{\parbox{0.95\textwidth}{
\textbf{Budget-Aware GRPO Pipeline}\\
\small 
1. \textbf{Difficulty Estimation:} Sample $K$ completions per prompt from base model $\rightarrow$ Compute success rate $\hat{p}(x)$\\
2. \textbf{Subset Selection:} Apply policy (\hard{}, \easy{}, \midsel{}, \rand{}) $\rightarrow$ Select 10\% subset\\
3. \textbf{GRPO Training:} Fine-tune model on selected subset using group advantages\\
4. \textbf{Evaluation:} Test on held-out benchmark
}}
\caption{Schematic overview of our experimental protocol}
\label{fig:pipeline}
\end{figure}

\subsection{Training procedure}

We employ Group Relative Policy Optimization (GRPO) following the formulation introduced in DeepSeekMath \cite{Shao2024DeepSeekMath}. More details on the specifics of the GRPO algorithm and hyperparameters used can be found in the appendix.

Our experiments span two benchmark datasets that probe different aspects of model capabilities. GSM8K provides 7,473 grade-school math problems requiring multi-step reasoning and arithmetic \cite{Cobbe2021GSM8K}. The Tracking 7 Shuffled Objects task from BIG-Bench Hard contains 250 problems testing the ability to maintain state through sequences of object swaps \cite{Suzgun2023BBH}, which we randomly split 50\%/50\% into train/test. We hold the number of prompts, rollouts per prompt, group size, KL schedule, max length, and total RL update steps constant across policies. This avoids claims being confounded by extra tokens/updates. %Finally, we include a KEGG-style knowledge task with 2,000 multiple-choice questions about biological pathways to assess whether difficulty-based selection benefits factual recall tasks \cite{Kanehisa2000KEGG}.

We evaluate three models: Qwen3-4B with 4 billion parameters representing a smaller but efficient architecture, Qwen3-14B with 14 billion parameters as a medium-scale model, and Phi-4 with 14 billion parameters, representing a different model family. All models are initialized from instruction-tuned checkpoints, ensuring they have basic instruction-following capabilities before GRPO fine-tuning begins.

\section{Experiments and Results}

We present our main experimental findings comparing difficulty-based subset selection strategies for GRPO fine-tuning.
Our results consistently demonstrate that training on the hardest examples yields superior performance on reasoning tasks.%, while showing negligible differences on knowledge tasks.

\subsection{Hard examples drive reasoning improvements}

\begin{figure}[h]
\centering
\includegraphics[width=\textwidth]{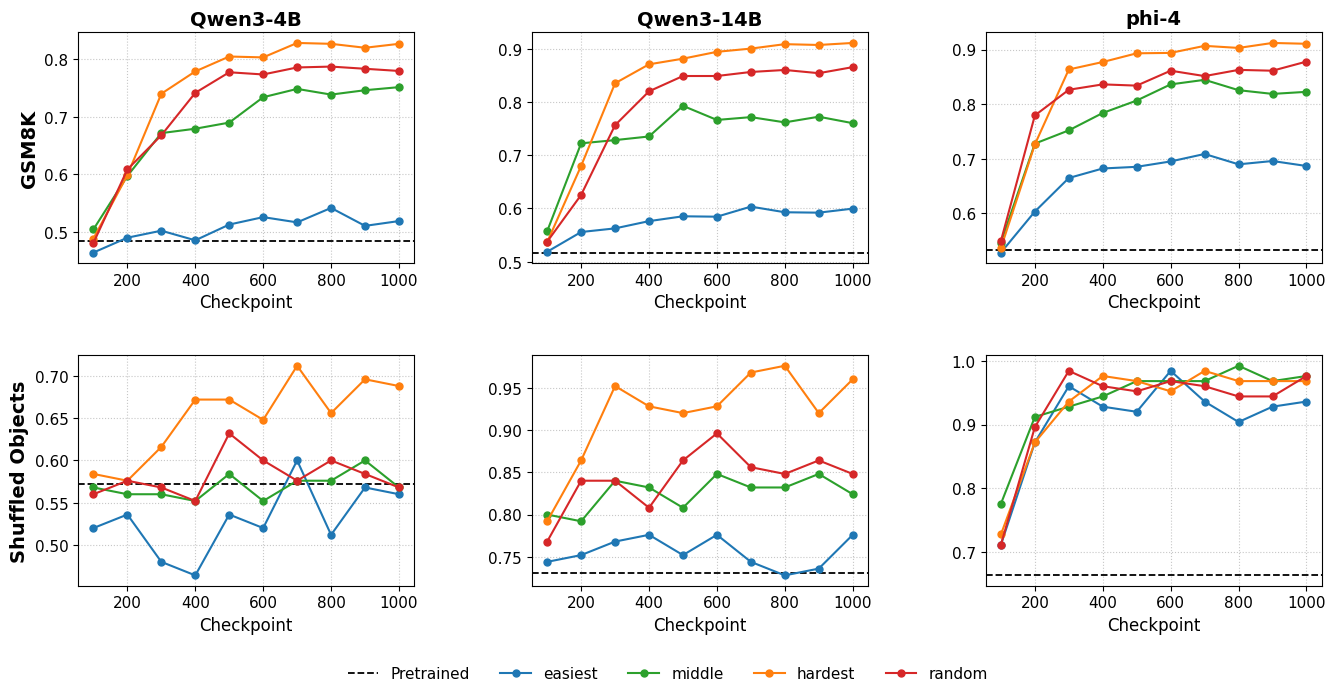}
\caption{GRPO training dynamics reveal early and persistent advantages of hard example selection. Each subplot shows test accuracy over 1000 training steps for models trained on different difficulty-based subsets (10\% of full data each). The hardest subset establishes superiority by step 300 and maintains this advantage.}

\label{fig:main_results}
\end{figure}

Figure~\ref{fig:main_results} and Table~\ref{tab:main_results}, summarize the absolute accuracy change (in percentage points) from the baseline after GRPO training, across selection policies, datasets, and models.

\begin{table}[h]
\centering

\label{tab:main_results}
\begin{tabular}{llcccc}
\toprule
\textbf{Dataset} & \textbf{Model} & \textbf{\easy} & \textbf{\midsel} & \textbf{\hard} & \textbf{\rand} \\
\midrule
\multirow{3}{*}{GSM8K} 
& Qwen3-4B  & 3.49 & 26.69 & \textbf{34.19} & \underline{29.49} \\
& Qwen3-14B & 8.26 & 24.34 & \textbf{39.42} & \underline{34.87} \\
& Phi-4     & 14.94 & 28.51 & \textbf{37.30} & \underline{36.16} \\
\midrule
\multirow{3}{*}{Shuffled Objects} 
& Qwen3-4B  & 0.80 & -3.20 & \textbf{13.60} & -1.60 \\
& Qwen3-14B & 3.20 & 10.12 & \textbf{22.40} & \underline{12.52} \\
& Phi-4     & \underline{29.60} & 28.80 & \textbf{32.80} & 28.80 \\
\end{tabular}
\caption{Absolute performance change (\%) after GRPO training with different subset selection policies compared to baseline. Bold indicates best performance, underline indicates second-best.}
\end{table}

The results reveal striking patterns in how difficulty-based selection impacts GRPO training effectiveness. On GSM8K, the superiority of hard example selection is unambiguous across all models. The \hard{} policy achieves remarkable performance improvements of 34.19\% for Qwen3-4B, 39.42\% for Qwen3-14B, and 37.3\% for Phi-4. These gains dwarf those achieved by easy selection, which manages only 3.49\%, 8.26\%, and 14.94\% respectively. The magnitude of these differences (exceeding 30 percentage points ) suggests that training on challenging examples fundamentally alters the learning dynamics under GRPO. We see training on either medium or random difficulties yield significant gains compared to training on easy examples, but not near the gains of training on the hardest subset - affirming the hypothesis that the harder the training subset, the larger the gains we see\cite{Cobbe2021GSM8K}.

The Tracking Shuffled Objects task amplifies these patterns. Hard selection yields positive gains of 13.60\%, 22.40\%, and 32.80\% for Qwen3-4B, Qwen3-14B, and Phi-4 respectively. For Qwen3-4B, we see all other subsets produce either zero or negative gains. For Qwen3-14B, improvements show a stepwise pattern: training on the easy subset performs poorly, medium and random subsets provide moderate gains, and the hard subset delivers the largest gains. For Phi-4, training on the hardest subset gives the highest gains, but training on any subset still produces substantial improvements, effectively saturating the benchmark \cite{Suzgun2023BBH}. 

% In stark contrast, the KEGG knowledge task shows no consistent benefit from any selection strategy, with performance changes scattered between -3\% and +9\% across models and policies. The absence of a clear pattern (\midsel{} performs best for Qwen3-4B at 9.45\% while all strategies show slight decrements for Qwen3-14B) confirms our hypothesis that factual knowledge cannot be effectively learned through GRPO's self-play mechanism regardless of example selection \cite{Kanehisa2000KEGG}. This null result serves as an important control, demonstrating that the dramatic advantages of hard selection on reasoning tasks stem from task-specific learning dynamics rather than general training artifacts.

% The influence of model families on selection dynamics provides additional insights. Qwen models exhibit extreme sensitivity to selection strategy, with performance gaps exceeding 30 percentage points between hard and easy selection. Llama3.1-8B shows more moderate differences, maintaining reasonable performance across all non-easy strategies while still benefiting most from hard examples. These architectural differences suggest that some models may be inherently more robust to suboptimal training data selection, though all benefit from principled difficulty-based curation.

\subsection{Performance dynamics during training}

Figure~\ref{fig:main_results} reveals how selection strategies shape learning trajectories from the earliest stages of training. The advantage of hard example selection manifests rapidly and decisively: by checkpoint 300, models trained on the hardest subset have already established clear superiority over all other selection strategies. This early divergence is particularly striking given that it represents less than a third of the total training steps, yet the performance ordering established at this point persists and often amplifies through the remainder of training. The consistency of this pattern across all three models suggests that hard examples provide fundamentally different learning signals that reshape the optimization landscape from the outset.

%Llama3.1-8B exhibits somewhat different dynamics while still confirming the superiority of hard selection. The easiest subset performs significantly worse throughout training, never recovering from its poor initial trajectory. Meanwhile, hard selection maintains marginal but consistent advantages over medium and random strategies. The more compressed performance range for Llama3.1-8B compared to Qwen models may reflect architectural differences in how these models process and learn from challenging examples. Crucially, across all models, the performance differences that emerge early in training show remarkable stability: we observe no cases where an initially inferior selection strategy overtakes a superior one later in training, suggesting that the initial subset choice creates path dependencies that fundamentally shape the final model capabilities.

\section{Analysis}

In this section, we try to better understand why hard examples are most effective for GRPO training in three ways. First, we examine learning dynamics through the lens of new metrics - ``learnable percentage". Second, we inspect performance on an out-of-distribution set to measure generalization capabilities when trained on different subsets. Third, we look at training just on examples where the base model is wrong vs where it is right.

\subsection{Why hard examples help: maintaining learnable samples}

A key insight into GRPO's preference for hard examples comes from tracking what fraction of training examples maintain non-zero outcome variance during training.
Recall that GRPO learns by contrasting outputs within a group; when all outputs in a group produce identical rewards (aka standard deviation of reward is zero), the advantages become zero and no learning occurs.

% We define the \textit{learnable percentage} over the course of training as the percentage of steps where the standard deviation is greater than zero (and therefore the advantage is nonzero):

% \[
% \text{\% Learnable} =
% \frac{1}{T} \sum_{t=1}^T 
% \mathbf{1}\!\left[ \mathrm{StdDev}(\mathbf{r}_t) > 0 \right] \times 100\%
% \]

% where $r_t$ is the reward for the $t$-th sample from $\pi_t$ during training.

\noindent \textbf{Learnable percentage.}
Let $T$ be the number of training updates. At update $t$, let
$\mathbf{r}_t = (r_{t,1},\ldots,r_{t,G}) \in \{0,1\}^G$
denote the vector of outcome rewards for the $G$ rollouts in the GRPO group under $\pi_t$.
We define the \emph{learnable percentage} as
\[
\text{Learnable \%}
\;=\;
100 \cdot \frac{1}{T}\sum_{t=1}^{T}
\mathbf{1}\!\left\{\,\mathsf{StdDev}(\mathbf{r}_t) > 0\,\right\},
\]
i.e., the fraction of training steps whose within-group reward standard deviation is nonzero (and therefore advantage is nonzero).

Table~\ref{tab:percent_learnable} shows the \% learnable across strategies and datasets. We see that, unsurprisingly, the hardest strategies have the highest \% learnable examples, meaning training on this data subset gives the most opportunity for the model to learn during training. Conversely, we see that the easy strategy has very few learnable examples. To test whether this effect depends on the model's baseline ability, we evaluate Llama-3.1-8B-Instruct \cite{dubey2024llama}, which has much lower initial performance on GSM8K. With a weaker baseline, we expect smaller performance gaps between subsets. Table~\ref{tab:llama} shows that while the hardest subset still achieves the largest gain, it is closer to the other subsets. Notably, all subsets have much higher \% learnable values than the corresponding Qwen results (e.g., for the easiest subset, $21\%$ vs.\ $4\%$ for Qwen3-4B).

\begin{table}[h]
\centering
\begin{tabular}{lcc|cc|cc}
\hline
 & \multicolumn{2}{c|}{Qwen3-4B} & \multicolumn{2}{c|}{Qwen3-14B} & \multicolumn{2}{c}{Phi-4} \\
Strategy & shuffleobj & gsm8k & shuffleobj & gsm8k & shuffleobj & gsm8k \\
\hline
Easy & 9.00 & 3.70 & 2.30 & 7.80 & 14.20 & 6.60 \\
Medium & 5.70 & \underline{24.50} & 7.20 & \underline{18.90} & 15.30 & \underline{23.30} \\
Hard & \textbf{14.40} & \textbf{34.10} & \textbf{13.50} & \textbf{29.50} & \underline{19.10} & \textbf{40.20} \\
Random & \underline{11.50} & 19.00 & \underline{8.00} & 18.40 & \textbf{19.30} & 21.50 \\
\hline
\end{tabular}
\caption{\% Learnable across models, strategies and datasets}
\label{tab:percent_learnable}
\end{table}

\begin{table}[h]
\centering
\begin{tabular}{l|c|c}
\hline
Strategy & Improvement over Base & \% Learnable \\
\hline
Easy & 36.54 & 20.90 \\
Medium & 39.42 & 42.00 \\
Hard & \textbf{40.33} & \textbf{55.20} \\
Random & 39.35 & 39.30 \\
\hline
\end{tabular}
\caption{Absolute accuracy improvement and \% learnable for Llama-3.1-8B-Instruct on GSM8K dataset}
\label{tab:llama}
\end{table}

\begin{figure}[H]
\centering
\includegraphics[width=0.6\textwidth]{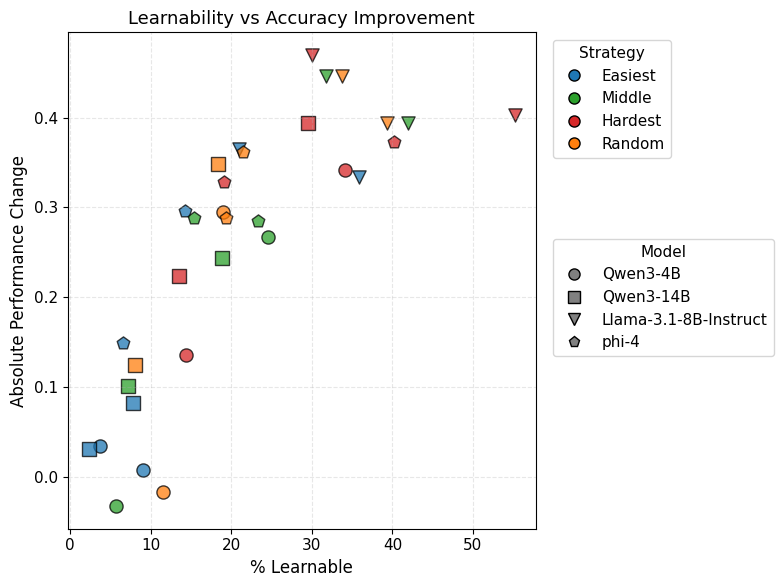}
\caption{Scatter plot showing the relationship between the percentage of learnable training examples and the resulting absolute improvement in model performance, across strategies and models. Colors indicate the strategy and marker shapes indicate the model. We see a strong positive correlation ($R^2=0.66$), indicating that performance improves with more learnable examples.}
\label{fig:learn_corr}
\end{figure}

Figure~\ref{fig:learn_corr} plots the relationship between the \% learnable and performance improvement, across all the models and strategies. We see a strong positive correlation, confirming that the underlying reason the harder subsets give better results is because they maintain more learnable examples throughout training.

Our analysis reveals why hard example selection proves dramatically more effective for GRPO training. Hard examples continue providing useful learning signal throughout training, while easy examples quickly become ``solved'' and stop contributing to improvement. This sustained learning opportunity explains the remarkable performance gaps we observe (often exceeding 30 percentage points between hard and easy selection strategies).

\subsection{Out-Of-Distribution Performance}

Having shown that harder training subsets improve test-set performance, we further evaluate on an out-of-distribution (OOD) test set. Specifically, we take the Qwen3-4B models trained on each subset and evaluate on AIME2025-I, a dataset of 15 questions significantly more difficulty than GSM8K \cite{matharena, opencompass_AIME2025}. Table~\ref{tab:AIME} presents the results.

\begin{table}[h]
\centering
\begin{tabular}{l c}
\toprule
\textbf{Strategy} & \textbf{Pass@8} \\
\midrule
Base     & 33.3\% \\
\specialrule{1.2pt}{0pt}{0pt}
Easy  & 33.3\% \\
Medium   & 26.7\% \\
Hard  & \textbf{40.0\%} \\
Random   & 33.3\% \\
\bottomrule
\end{tabular}
\caption{Pass@8 on AIME-2025 of Qwen3-4B trained on different subsets}
\label{tab:AIME}
\end{table}

We see that the training on the hardest subset is the only one to outperform the base Qwen3-4B model. Surprisingly, training on the medium subset actually does worse than training on easy or random subsets. 

We see that training on the hardest subset is the only condition that meaningfully outperforms the base Qwen3-4B model, achieving a relative improvement of +20\%. This reinforces the trend observed on the in-distribution test set: exposure to more challenging problems during training generalizes better to harder OOD problems. Interestingly, overall performance on the easy and random subsets remains identical to the base model, with closer inspection showing that these subsets improved performance on certain problems but regressed performance in other areas. In contrast, training on the medium difficulty subset yields a notable drop in performance, with no other gains.

Overall, these results show that the benefits of hard-example training extend beyond the original distribution, improving robustness to unseen, more challenging problems.

\subsection{Base Wrong vs Base Right: Decomposing GRPO Value}

Having established that hard examples (those with low pass@k success rates) drive the most value in GRPO training, we now validate this principle from a complementary perspective. We investigate whether the superiority of hard examples stems from a more fundamental property: that they represent problems the base model cannot reliably solve. %This analysis tests whether ``hardness'' as measured by pass@k rates and ``hardness'' as measured by base model failure are two facets of the same underlying phenomenon.

We categorize training examples into four types based on base model performance (here base means the model pre GRPO training):

\begin{enumerate}
\item \textbf{Base Wrong}: Examples the base model gets wrong (accuracy \textless  25\% )
\item \textbf{Base Right}: Examples the base model gets right (accuracy $\geq$ 25\% )
\item \textbf{Base Right, Same Size As Base Wrong}: A subset of the full Base Right Set that is the same size as Base Wrong, to account for model size
\item \textbf{All}: All the available examples in the train set
\end{enumerate}

We hypothesize that training just on the examples where the base model is wrong will significantly match the base right subsets, as it provides the most opportunity for genuine capability improvement and higher learnable percentage.

\subsubsection{Results}

Table~\ref{tab:base_wrong_right} presents the absolute performance after training on each subset:

\begin{table}[h]
\centering
% Column width options:
% - p{Xcm}: Fixed width paragraph column (text wraps if needed)
% - c: Centered column (auto-width based on content)
% - l: Left-aligned column (auto-width)
% - r: Right-aligned column (auto-width)
% Example: Change p{3.5cm} to p{4cm} for wider first column
% Example: Change p{1.2cm} to c for auto-width centered columns
\begin{tabular}{p{3.5cm}|ccc|ccc}
\toprule
& \multicolumn{3}{c|}{ShuffleObj} & \multicolumn{3}{c}{GSM8K} \\
Training Set & Phi-4 & Qwen3-4B & Qwen3-8B & Phi-4 & Qwen3-4B & Qwen3-8B \\
\midrule
Base Wrong & \textbf{0.98} & \textbf{0.73} & 0.81 & \textbf{0.92} & \textbf{0.84} & \textbf{0.86} \\
%Random (Same Size) & \textbf{0.98} & 0.69 & 0.80 & 0.87 & 0.74 & \textbf{0.86} \\
Base Right & \textbf{0.98} & 0.64 & 0.83 & 0.80 & 0.70 & 0.72 \\
Base Right (Same Size) & \textbf{0.98} & 0.67 & 0.75 & 0.78 & 0.68 & 0.71 \\
All & 0.97 & 0.66 & \textbf{0.84} & 0.85 & 0.77 & 0.85 \\
\bottomrule
\end{tabular}
\caption{Absolute performance on test sets after training on different subsets based on base model performance. Bold indicates best or tied-best performance.}
\label{tab:base_wrong_right}
\end{table}

Table~\ref{tab:base_wrong_improvements} shows the relative improvements of Base Wrong training over both Base Right subsets. Table~\ref{tab:base_wrong_size} shows the sizes of each subset - Base Wrong is, on average, less than half the size of Base Right. 

\begin{table}[h]
\centering
% Column width options:
% - p{Xcm}: Fixed width paragraph column (text wraps if needed)
% - c: Centered column (auto-width based on content)
% - l: Left-aligned column (auto-width)
% - r: Right-aligned column (auto-width)
% Example: Change p{2.5cm} to p{3cm} for wider first column
% Example: Change c to p{1.2cm} for fixed-width columns
\begin{tabular}{p{2.5cm}|p{1.1cm}p{1.1cm}p{1.1cm}|p{1.1cm}p{1.1cm}p{1.1cm}|c}
\toprule
& \multicolumn{3}{c|}{ShuffleObj} & \multicolumn{3}{c|}{GSM8K} & \\
Metric & Phi-4 & Qwen3-4B & Qwen3-8B & Phi-4 & Qwen3-4B & Qwen3-8B & Average \\
\midrule
\multicolumn{8}{l}{\textit{Improvement Over Base Right (Same Size)}} \\
Absolute & 0.0\% & 6.0\% & 5.8\% & 14.0\% & 16.0\% & 15.0\% & 10.2\% \\
Relative & 0.0\% & 9.0\% & 7.7\% & 17.9\% & 23.5\% & 21.1\% & 14.3\% \\
\midrule
\multicolumn{8}{l}{\textit{Improvement Over Base Right (Full)}} \\
Absolute & 0.0\% & 9.0\% & -2.0\% & 12.0\% & 14.0\% & 14.0\% & 7.8\% \\
Relative & 0.0\% & 14.1\% & -2.4\% & 15.0\% & 20.0\% & 19.4\% & 11.0\% \\
\bottomrule
\end{tabular}
\caption{Improvements from training on Base Wrong examples compared to Base Right examples. Relative improvement calculated as (Base Wrong - Base Right) / Base Right.}
\label{tab:base_wrong_improvements}
\end{table}

\subsubsection{Analysis}

The results strongly support our hypothesis that GRPO learning value concentrates in examples the base model gets wrong:

\textbf{Base Wrong consistently outperforms Base Right}: Across models and tasks, training exclusively on Base Wrong examples achieves equal or superior performance to training on Base Right, even when Base Right has significantly more examples. The average relative improvement of 14.3\% over size-matched Base Right subsets demonstrates that these examples provide fundamentally different learning signals. Even when we train on the full Base Right set, we see that Base Wrong outperforms it by 11\% on average.

\textbf{Base Wrong matches / outperforms All}: Training on all Base Wrong samples actually slightly outperforms training on All samples on average (85\% vs 82\%), further supporting that most, if not all, meaningful training signal is concentrated in the Base Wrong subset.

\textbf{Task-specific patterns emerge}: On GSM8K, the advantage is particularly pronounced, with Base Wrong training achieving 15-23.5\% relative improvements. This aligns with our earlier findings that mathematical reasoning benefits most from challenging examples. ShuffleObj shows more modest gains, potentially due to the discrete nature of object tracking limiting the diversity of solution paths.

\textbf{Model-specific sensitivity}: Phi-4 shows less differentiation between strategies on ShuffleObj (all achieving 0.98), suggesting potential ceiling effects. However, on GSM8K, even Phi-4 shows substantial 17.9\% relative improvement, confirming the generality of the phenomenon.

\textbf{Practical implications}: These findings provide a practical recipe for GRPO dataset curation: practitioners can identify valuable training examples by simply evaluating base model performance, without needing to know which examples will ultimately improve. This avoids the circular dependency inherent in selecting ``learnable'' examples and provides a straightforward criterion for data collection.

\begin{table}[h]
\centering
\begin{tabular}{p{3.5cm}|ccc|ccc}
\toprule
& \multicolumn{3}{c|}{ShuffleObj} & \multicolumn{3}{c}{GSM8K} \\
Training Set & Phi-4 & Qwen3-4B & Qwen3-8B & Phi-4 & Qwen3-4B & Qwen3-8B \\
\midrule
Base Wrong & 17 & 37 & 13 & 343 & 409 & 371 \\
Base Right & 108 & 88 & 112 & 657 & 591 & 629 \\
All & 125 & 125 & 125 & 1000 & 1000 & 1000 \\
\bottomrule
\end{tabular}
\caption{Subset sizes for each dataset, model, and training strategy. On average, Base Wrong is only 42\% the size of Base Right, and yet consistently significantly outperforms this subset. Note that GSM8K is clipped to 1000 since we only run for 1000 steps with 1 prompt per update.}
\label{tab:base_wrong_size}
\end{table}

The concentration of learning value in Base Wrong examples validates our central thesis from a complementary angle. Our two experimental approaches converge on the same insight: examples with low pass@k rates (our ``hardest'' selection) are predominantly those where the base model fails (``base wrong''), while examples with high pass@k rates (our ``easy'' selection) are those the base model already masters (``base right''). This convergence demonstrates that whether we measure difficulty through multi-sample success rates or binary base model performance, we identify the same high-value training examples: those at the frontier of model capabilities where learning can actually occur.

\section{Conclusion}

This paper investigated a critical question for resource-constrained language model fine-tuning: under a fixed budget for example selection, which difficulty-based strategy maximizes the effectiveness of GRPO? Through comprehensive experiments across multiple models and tasks, we established that training on the hardest examples consistently yields superior performance on reasoning benchmarks. 

We validated this ``hard examples are all you need'' principle from two complementary perspectives. First, our difficulty-based selection experiments showed that examples with the lowest pass@k success rates drive performance gains up to 47\%, while easy examples yield minimal improvements. Second, our analysis of training data categorized by base model performance revealed that nearly all GRPO learning value concentrates in examples the base model initially fails to solve. Training exclusively on examples where the base model achieves low success rates ($\hat{p}(x) < 0.25$) yields up to 23.5\% relative improvements compared to training on examples the base model already solves correctly ($\hat{p}(x) \geq 0.25$). 

This convergence of evidence from both perspectives confirms a fundamental insight: GRPO learning requires training at the frontier of model capabilities. Examples the base model already masters reliably (high pass@k or those it solves correctly) are effectively the easiest and contribute minimal learning value, while examples where the model struggles (low pass@k or those it fails to solve) provide the variance in outcomes necessary for GRPO's contrastive learning mechanism. This unified understanding provides practitioners with multiple practical approaches for identifying high-value training data while avoiding circular dependencies in dataset curation.

\bibliography{references}
\bibliographystyle{iclr2026_conference}

% Appendix section
\appendix
\section{Related Work}

Our work intersects several research areas: reinforcement learning from human feedback, example selection and curriculum learning, reasoning benchmarks, and knowledge evaluation tasks.
We review each area and position our contributions within this broader context.

\subsection{Reinforcement Learning from Human Feedback and GRPO}

Reinforcement learning from human feedback (RLHF) has emerged as a dominant paradigm for aligning language models with human preferences \cite{Christiano2017DRLHF,Ouyang2022InstructGPT}.
The standard RLHF pipeline involves training a reward model from human preferences, then optimizing the language model policy using algorithms like Proximal Policy Optimization (PPO) \cite{Schulman2017PPO}.
However, PPO-based RLHF suffers from high computational costs, training instability, and significant variance in gradient estimates.

Group Relative Policy Optimization (GRPO), introduced in DeepSeekMath \cite{Shao2024DeepSeekMath}, addresses these challenges by reformulating the optimization problem.
Instead of training a separate reward model, GRPO directly optimizes using group-relative advantages computed from multiple samples per prompt.
For each training prompt, GRPO samples a group of $G$ outputs, computes their rewards (e.g., correctness on math problems), and uses the deviation from the group mean as the advantage signal.
This approach reduces variance through the group baseline while eliminating the need for reward model training.

Our work builds directly on GRPO's formulation but introduces a critical new dimension: budget-aware example selection.
While previous GRPO applications assume uniform sampling from the training distribution, we show that strategic selection based on example difficulty can substantially improve learning efficiency, particularly relevant when data acquisition costs constrain the training set size.

\subsection{Example Difficulty and Data Selection}

The importance of example selection in machine learning has long been recognized, with multiple research threads addressing different aspects of this problem.

\textbf{Curriculum learning} suggests that training on examples in order of increasing difficulty can improve convergence and final performance \cite{Hacohen2019Curriculum}.
However, curriculum learning typically assumes access to the full dataset and focuses on presentation order rather than subset selection.
Our work differs by operating under a strict budget constraint where only a fraction of examples can be used.

\textbf{Data valuation and influence} methods aim to identify the most valuable training examples.
Data Shapley \cite{Ghorbani2019DataShapley} uses cooperative game theory to assign value to each training example based on its marginal contribution.
However, these methods are computationally expensive and designed for supervised learning rather than RLHF settings.

\textbf{Coreset selection} techniques identify representative subsets that approximate the full dataset's gradient \cite{Killamsetty2021GradMatch,Killamsetty2021GLISTER}.
GradMatch selects examples whose gradients best match the full dataset gradient, while GLISTER focuses on generalization-based selection.
These methods provide theoretical guarantees but require access to labels and gradient computation that may not be available in our budget-constrained setting.

\textbf{Example forgetting and memorization} studies have shown that neural networks exhibit consistent patterns in which examples are learned or forgotten during training \cite{Toneva2019Forgetting}.

\textbf{Synthetic data} is often an area of investigation for training, especially since it is often easier to control for difficulty. Our work doesn't explicitly investigate using synthetic data and instead selects from a real, static larger dataset \cite{condition, deepdistill, sparq}. 

``Unforgettable'' examples are learned early and retained, while others are repeatedly learned and forgotten.
The Data Diet work \cite{Paul2021DataDiet} leverages Error L2-Norm (EL2N) scores computed early in training to identify important examples, showing that networks can be trained on small subsets without performance loss.

Our approach differs from these prior works in several ways:
(1) We operate in the RLHF/GRPO setting where rewards are binary and computed at inference time;
(2) We use the base model's multi-sample success rate as a difficulty proxy, requiring no training to compute;
(3) We explicitly compare different difficulty-based selection strategies under identical budgets.

\subsection{Reasoning Benchmarks and Evaluation}

Evaluating reasoning capabilities in language models requires carefully designed benchmarks that test different aspects of logical and mathematical thinking.

\textbf{GSM8K} \cite{Cobbe2021GSM8K} has become the standard benchmark for grade-school mathematical reasoning, containing 8,792 problems requiring multi-step arithmetic and logical reasoning.
The dataset's problems are linguistically diverse but mathematically elementary, making it ideal for studying reasoning without confounding advanced mathematical knowledge.

\textbf{BIG-Bench Hard (BBH)} \cite{Suzgun2023BBH} curates 23 challenging tasks from the broader BIG-Bench suite where language models initially showed poor performance.
The Tracking Shuffled Objects task, which we use in our experiments, requires models to track entity positions through a series of swapping operations: a pure reasoning task with minimal knowledge requirements.

\textbf{Chain-of-thought prompting and self-consistency} \cite{Wang2022SelfConsistency} have proven effective for improving reasoning performance.
Self-consistency samples multiple reasoning paths and aggregates answers, often yielding substantial accuracy improvements.
We leverage this insight in our difficulty estimation, using multi-sample success rates to robustly measure example hardness.

%\subsection{Knowledge Evaluation Tasks}

% While reasoning tasks test procedural capabilities, knowledge tasks evaluate factual recall and understanding.

% \textbf{KEGG (Kyoto Encyclopedia of Genes and Genomes)} \cite{Kanehisa2000KEGG} provides a comprehensive database of biological pathways and systems.
% While originally designed for bioinformatics research, KEGG's structured knowledge makes it valuable for constructing factual question-answering tasks that test specific domain knowledge rather than reasoning ability.

% Our inclusion of a KEGG-style knowledge task serves as a critical control: we hypothesize that such tasks, which require static factual knowledge rather than procedural reasoning, will show different selection dynamics under GRPO.
% This comparison helps delineate the boundaries of when difficulty-based selection provides benefits.

\subsection{GRPO}
The GRPO algorithm operates by sampling groups of outputs for each training prompt and using relative performance within each group to compute advantages. Specifically, for each prompt $q$ in the selected subset, the algorithm samples $G$ outputs $\{o_1, ..., o_G\}$ from the current policy $\pi_\theta$, computes binary rewards $r_i \in \{0, 1\}$ based on correctness, and calculates a group baseline $\bar{r} = \frac{1}{G}\sum_{i=1}^G r_i$. The advantage for each output is then computed as $A_i = r_i - \bar{r}$, which naturally centers the learning signal around the group's average performance. These advantages drive policy updates while KL regularization to a reference policy $\pi_{\text{ref}}$ prevents catastrophic distribution shifts.

The training objective maximizes the expected advantage-weighted log-likelihood while minimizing KL divergence from the reference policy:
\[
\mathcal{L}_{\text{GRPO}} = -\mathbb{E}_{q \sim S} \mathbb{E}_{o \sim \pi_\theta(\cdot|q)} \left[ A(o, q) \cdot \log \pi_\theta(o|q) - \beta \cdot \text{KL}(\pi_\theta || \pi_{\text{ref}}) \right]
\]
where $\beta$ controls the strength of KL regularization.

\subsection{Closest GRPO Work}

Recent work has explored various aspects of GRPO optimization that relate closely to our budget-aware selection study:

\textbf{Difficulty-targeted selection and replay.}
Sun et al. propose a data-efficiency recipe for GRPO-style RL fine-tuning that (i) targets adaptive difficulty online, prioritizing questions of moderate difficulty expected to yield informative gradients, and (ii) introduces rollout replay to reuse recent samples and lower per-step cost. Across six LLM-dataset pairs they report 25-65\% faster time-to-target performance than vanilla GRPO, suggesting that what you train on (and when) can matter as much as how much you train. Our paper shares the focus on difficulty-aware allocation under constraints, but differs in scope: we study budgeted, offline subset selection (hard vs. medium vs. easy vs. random) with a fixed labeling budget and analyze why ``hardest'' can dominate on reasoning tasks, whereas Sun et al. optimize online time-to-quality with an adaptive, typically moderate-difficulty target and replay \cite{Sun2025DiffSelReplayGRPO}.

\textbf{Process rewards compatible with GRPO.}
Cui et al. (PRIME) tackle the sparsity and credit-assignment limits of outcome-only rewards by learning implicit process reward models online from outcome labels alone, then combining dense token-level signals with outcome rewards during RL. PRIME is compatible with standard advantage formulations (including GRPO) and removes a separate reward-modeling stage; empirically, they show sizable gains on competitive math and coding (e.g., ~15\% average over SFT with a 7B base). Our study focuses on which examples to buy under an outcome-reward GRPO setup; PRIME is complementary, and our framework could be extended to ask whether difficulty-aware selection interacts with dense process rewards in similar ways (e.g., whether ``hardest'' remains optimal when per-step feedback reduces variance) \cite{Cui2025PRIME}.

\textbf{GRPO biases and token efficiency.}
Liu et al. present a critical analysis of R1-Zero-like training and identify an optimization bias in GRPO: the objective can inflate response length, particularly for incorrect outputs, harming token efficiency. They introduce Dr. GRPO, an unbiased variant that preserves reasoning quality while improving efficiency, and demonstrate strong AIME-24 results with a 7B base model. For our budget-aware study, their findings motivate length-controlled evaluations and guardrails when comparing difficulty-conditioned subsets (e.g., ensuring that harder-subset gains are not confounded by pathological length growth and considering Dr. GRPO as a robustness check) \cite{Liu2025DrGRPO}.

\textbf{Positioning.}
Together, these works closest to GRPO underscore three levers we explicitly separate and test: (a) example selection by difficulty (Sun et al.), (b) reward shaping via dense process feedback (Cui et al.), and (c) objective design to avoid length bias (Liu et al.). Our contribution is orthogonal and complementary: a budget-aware, offline selection protocol and theory-informed analysis showing why, under outcome-level GRPO, training on the hardest decile can sustain a higher fraction of ``learnable'' examples (non-degenerate group variance) and thus larger gains on reasoning tasks, while we adopt length controls in line with Dr. GRPO insights \cite{Sun2025DiffSelReplayGRPO,Cui2025PRIME,Liu2025DrGRPO}.

% \subsection{Out-Of-Distribution}

% Recent research has increasingly focused on evaluating large language models (LLMs) in out-of-distribution (OOD) scenarios, highlighting their potential and limitations in such contexts. Previous work conducted an empirical study on LLaMA models ranging from 7B to 65B parameters, demonstrating that simple cosine distance-based OOD detectors outperform traditional methods. Another work introduced LLM-GOOD, a framework that integrates LLMs' zero-shot capabilities with graph neural networks to efficiently detect OOD nodes in text-attributed graphs, thereby reducing the need for extensive human annotation. Additionally,  recent works have investigated the performance of reward models in OOD settings, as well as more traditional OOD evaluation \cite{llmood, rewardmodels, oodenergy, oodgraph, oodsurvey}. These studies collectively underscore the evolving landscape of LLM evaluation in OOD contexts, emphasizing the need for tailored approaches to assess model reliability beyond standard benchmarks

\subsection{Positioning Our Contributions}

Our work makes several distinct contributions relative to this prior literature:

\begin{enumerate}
\item We are the first to systematically study budget-aware example selection specifically for GRPO fine-tuning, addressing a practical constraint faced by practitioners.

\item We provide a unified evaluation framework comparing difficulty-based selection strategies across multiple task types and model scales.

\item We offer theoretical insight connecting GRPO's group-advantage mechanism to the effectiveness of hard-example selection, grounded in variance reduction principles from reinforcement learning theory.

\item We demonstrate that selection strategy effectiveness depends critically on task type, with reasoning tasks benefiting from hard examples while knowledge tasks show no such preference.
\end{enumerate}

These contributions provide both theoretical understanding and practical guidance for efficient RLHF fine-tuning under resource constraints.
\section{Hyperparameters}

We set the group size $G = 8$ for our main experiments based on preliminary studies showing this balances variance reduction with computational efficiency. The KL coefficient $\beta = 0.1$ provides sufficient regularization without overly constraining learning. Training proceeds for 1000 steps with a learning rate of $3 \times 10^{-5}$ using cosine decay, processing 1 prompt per gradient update due to memory constraints. We use 4-bit quantized models from Unsloth \cite{unsloth}. We evaluate model performance every 100 steps to track learning dynamics, using the initial SFT checkpoint as our reference policy throughout training.
\section{LLM Use}

Large Language Models (LLMs) were used for the following purposes during this research:

\begin{enumerate}
\item \textbf{Writing some of the code for the experiments}: LLMs assisted in generating and debugging portions of the experimental code, including data processing scripts and evaluation utilities.

\item \textbf{Proofreading the text in the paper}: LLMs were used to identify grammatical errors, improve clarity, and ensure consistency in the manuscript.

\item \textbf{LaTeX support}: LLMs provided assistance with LaTeX syntax for complex table formatting and figure placement.
\end{enumerate}

\end{document}